\documentclass[letterpaper]{article} 
\usepackage[submission]{aaai23}  
\usepackage{times}  
\usepackage{helvet}  
\usepackage{courier}  
\usepackage[hyphens]{url}  
\usepackage{graphicx} 
\usepackage{natbib}  
\usepackage{caption} 
\usepackage{algorithm}
\usepackage{algorithmic}
\usepackage{multirow}
\usepackage{amsmath}

\usepackage[subfigure]{tocloft}
\usepackage{subfigure}
\usepackage{amssymb}

\usepackage{newfloat}
\usepackage{listings}
\DeclareCaptionStyle{ruled}{labelfont=normalfont,labelsep=colon,strut=off} 
\floatstyle{ruled}
\newfloat{listing}{tb}{lst}{}
\floatname{listing}{Listing}
\title{Human Pose Driven Object Effects Recommendation}
\author{
    Zhaoxin Fan\equalcontrib \textsuperscript{\rm 1}, Fengxin Li\equalcontrib \textsuperscript{\rm 1}, Hongyan Liu\textsuperscript{\rm 2}, \\Jun He \textsuperscript{\rm 1}, Xiaoyong Du \textsuperscript{\rm 1}
}
\affiliations{
    \textsuperscript{\rm 1}Renmin university of china\\
    \textsuperscript{\rm 2}Tinghua university\\

    fanzhaoxin@ruc.edu.cn, lifengxin@ruc.edu.cn, hyliu@tsinghua.edu.cn, \\hejun@ruc.edu.cn, duyong@ruc.edu.cn
}

\begin{document}

\maketitle

\begin{abstract}
In this paper, we research the new topic of object effects recommendation in micro-video platforms, which is a challenging but important task for many practical applications such as advertisement insertion. To avoid the problem of introducing background bias caused by directly learning video content from image frames, we propose to utilize the meaningful body language hidden in 3D human pose for recommendation. To this end, in this work, a novel human pose driven object effects recommendation network termed PoseRec is introduced. PoseRec leverages the advantages of 3D human pose detection and learns information from multi-frame 3D human pose for video-item registration, resulting in high quality object effects recommendation performance. Moreover, to solve the inherent ambiguity and sparsity issues that exist in object effects recommendation, we further propose a novel item-aware implicit prototype learning module and a novel pose-aware transductive hard-negative mining module to better learn pose-item relationships. What's more, to benchmark methods for the new research topic, we build a new dataset for object effects recommendation named Pose-OBE. Extensive experiments on Pose-OBE demonstrate that our method can achieve superior performance than strong baselines.
\end{abstract}

\section{Instruction}

Personalized recommendation, an important solution to information overload, has attracted numerous attention in both academia and industry. Given both user information and item information, personalized recommendation tends to mine user preferences by examining the relationship between users and items, hence providing persuasive recommendation results. In the past decades, personalized recommendation has been leveraged to benefiting many practical applications, e.g., news spreading \cite{DBLP:conf/aaai/WuWWH021, DBLP:conf/www/WuWQLTLH0022}, goods selling \cite{DBLP:journals/ipm/ZhengLL21, DBLP:conf/wsdm/SingerRENGLHK22}, etc., producing great economical values. Most of the above applications are products of the traditional business model. Recently, with the increasing popularity of micro-video platforms, such as Tiktok and Kwai, applying personalized recommendation to micro-videos and live streaming becomes popular, bringing new business models and recommendation paradigms. 

\begin{figure*}[t]
  \centering
  \includegraphics[width=0.70\linewidth]{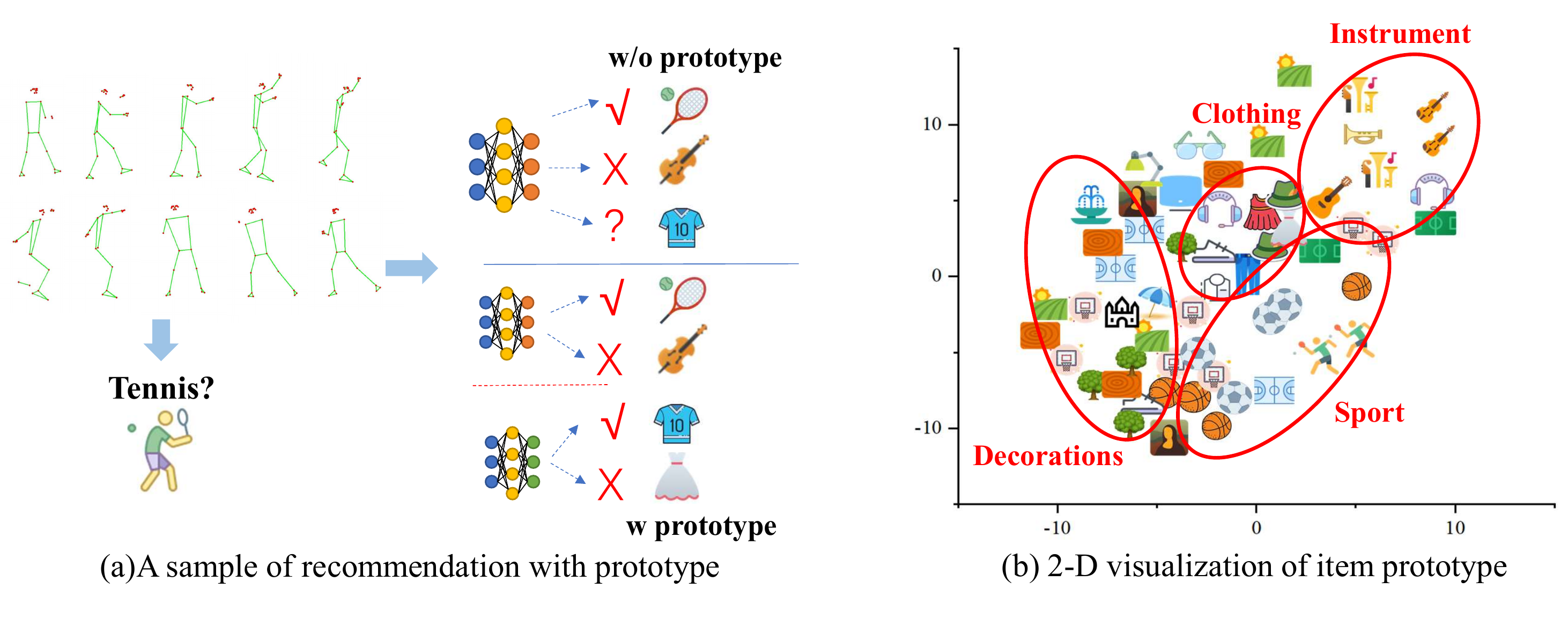}
  	\vspace{-0.15in}
    \caption{ Illustration of object effects recommendation. (a) An example of utilizing human pose for object effects recommendation.  (b) Distribution of learned prototypes and items.}
	\label{fig_sample}
	\vspace{-0.25in}
\end{figure*}

In the field of micro-video platforms, most of existing micro-videos and live streaming recommendation methods focus on recommending micro-videos/live streaming to users \cite{, DBLP:journals/tmm/CaiQFX22, DBLP:journals/kbs/CaoMRQN20, DBLP:journals/apin/ZhangLMHD22, DBLP:journals/mta/LinCCC21, DBLP:conf/icmcs/LiuXZC20} according to their preferences. For example, methods like Liu et al. \cite{DBLP:conf/www/LiuCLH19} propose a user-video co-attention network for the micro-video recommendation, which utilizes the attention mechanism to mine the relationship between users' preferences and videos.  Wei et al. \cite{DBLP:conf/mm/WeiWN0HC19} propose a multi-modal graph convolution to better leverage the multi-modal content information hidden images, audio, and text respectively. Yi et al.  \cite{DBLP:journals/corr/abs-2107-07268} propose a cross-modal variational auto-encoder for content-based micro-video background music recommendation. This kind of recommendation has achieved excellent performance and has significantly benefited the research community and many industrial companies. Nevertheless, we find another kind of recommendation, termed object effects recommendation, though equally important, attracts little research interest. 

In this paper, we research the very important but new topic of object effects recommendation for micro-videos. Given a video and a dataset of corresponding items, the topic aims at scoring and ranking these items so that the system can recommend items that are most relevant to the video content to the user. The items would be regarded as object effects and then be intelligently added into the micro-video to improve its quality, with the help of video edition technologies. This setting is very useful and widely applicable. For example, on the one hand, we can use the recommendation algorithm to add advertisements according to the video content. On the other hand, one can use the recommendation result to post-processing a micro-video.

To achieve the object effects recommendation goal, a very important aspect is how to extract video content. A straight-forward idea is to use a deep learning model to learn image/video-level features to represent the expected scene content. However, we find that since most micro-videos are human-centered, directly learning deep features from videos would introduce bias. More specifically, on the one hand, we hope to extract features that can best describe human behavior and action in the micro-video; on the other hand, the deep network tends to learn information about the background scenes. This would degrade the recommendation performance. Compared to learning video content directly, we observe that body languages hidden in human poses are very useful information that has long been neglected in recommendation. To this end, we propose a novel \emph{Human Pose Driven object Effects Recommendation Network} named PoseRec, which greatly leverages human poses for object effects recommendation in micro-videos. In our work, 3D human pose trajectories are extracted from videos and used to learn high level video contents. These contents represent the user preference well for recommendation by abstracting sequential body language. For example, as shown in Fig \ref{fig_sample} (a), once we extract a pose that a human was waving, the recommendation system would guess that the human in the video is playing tennis and the video is highly related to tennis, hence it would rank tennis balls and tennis shoes with higher scores for recommendation.

Though utilizing 3D human pose for object effects recommendation is interesting and superior to directly use video features. The topic arises new challenges. Specifically, there are two serious issues: inherent ambiguity and sparsity exist in pose-item registration. The former issue means that there is a multiplicity of solutions between a pose and a large number of fine-grained items. The latter issue means that since there are too many items, it is very hard to distinguish positive items and negative items, especially when hard negative samples are needed during network training. To solve the two issues, we further propose two novel modules named item-aware implicit prototype learning module and pose-aware transductive hard-negative mining module, respectively. The first module solves the ambiguity problem by implicitly clustering different items into prototypes, while the second module solves the sparsity problem by utilizing the pose-to-pose mapping to transductively sampling hard-negative samples during network training.

To the best of our knowledge, we are the first to research object effects recommendation in the field of the micro-video platform. To benchmark object effects recommendation methods, we build a novel dataset named Pose-OBE, consisting of 212 micro-videos. Each video is annotated with object effects that are most suitable for the scenario, by a micro-video operation specialist.  Each item (object effect) is tagged with a 9-dimensional description including name, usage, shape, color, et al. We conduct extensive experiments on Pose-OBE and compare our method with several strong baselines. Experimental results show that our method significantly outperforms baseline methods and can produce convincing recommendation results.

Our contribution can be summarized as: 1) We are the first to research object effects recommendation in micro-video platforms. A novel method named PoseRec is proposed to leverage body language hidden in 3D human poses for recommendation. 2) Two novel modules named item-aware implicit prototype learning module and pose-aware transductive hard-negative mining module are proposed to solve the inherent ambiguity and sparsity issues in human pose driven object effects recommendation. 3) A new object effects recommendation benchmark dataset named Pose-OBE is presented, along with extensive experiments on this dataset to demonstrate the superiority of PoseRec.



\begin{figure*}[t]
  \centering
  \includegraphics[width=0.75\linewidth]{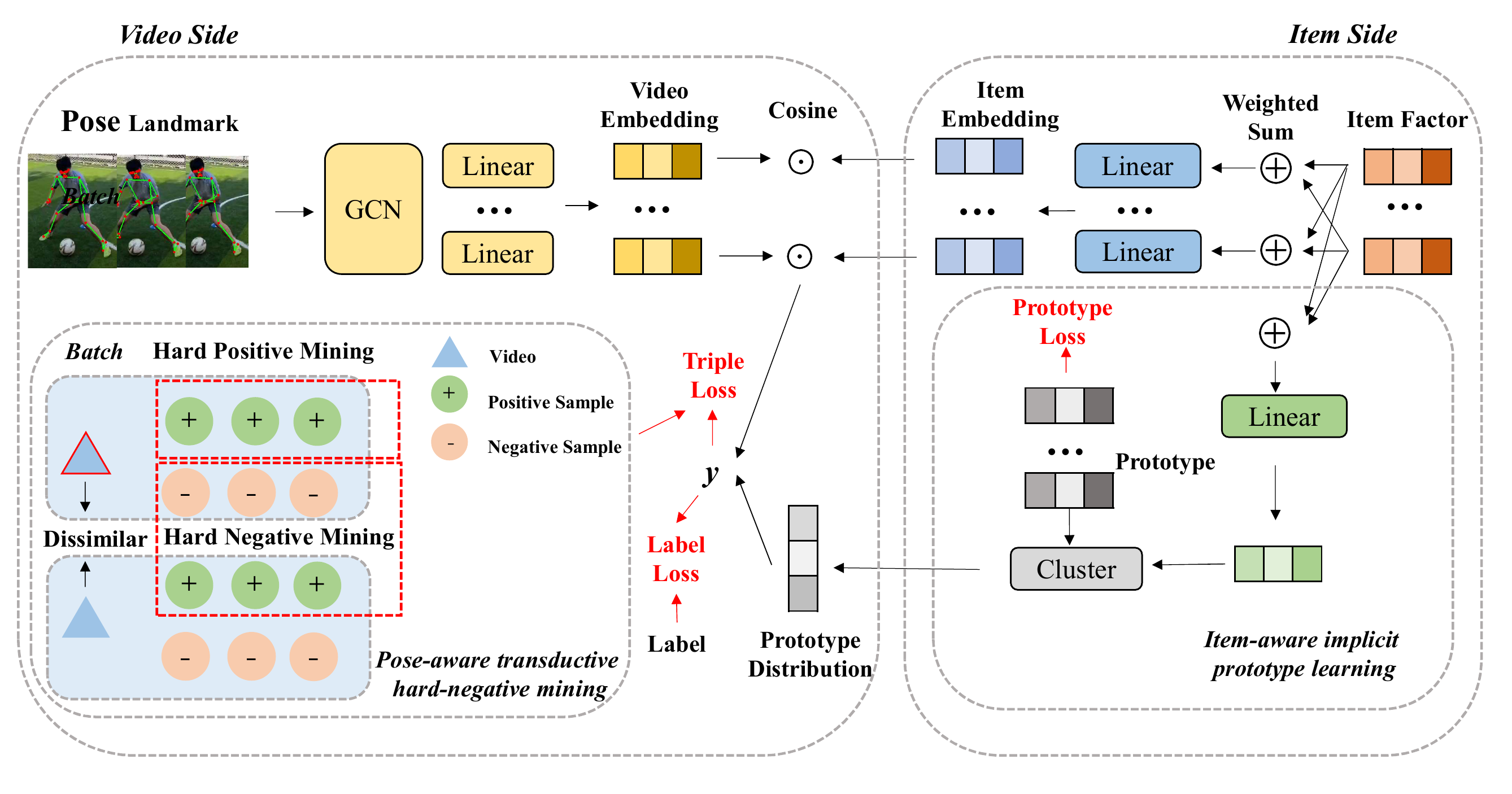}
  \vspace{-0.15in}
    \caption{Framework of Human Pose Driven Object Effects Recommendation. }
    \vspace{-0.25in}
	\label{fig_framework}
\end{figure*}

\section{Related work}

\noindent \textbf{Human pose estimation} Human pose estimation has attracted a lot of research interests in recent years \cite{DBLP:journals/tog/YiZ021, DBLP:journals/pr/BenzineLPA21, DBLP:conf/cvpr/XuT21, DBLP:conf/cvpr/LiXCBYL21, DBLP:conf/cvpr/GongZF21, DBLP:conf/cvpr/0007WSKS21}. In general, existing human pose estimation methods can be divided into two categories: bottom-up methods \cite{DBLP:conf/cvpr/CaoSWS17, DBLP:conf/eccv/KocabasKA18, DBLP:conf/cvpr/KreissBA19, DBLP:conf/cvpr/LiWZMFL19, DBLP:journals/ijcv/LiuSWCCA21} and top-down methods \cite{ DBLP:conf/iccv/FangXTL17, DBLP:conf/eccv/XiaoWW18, DBLP:conf/cvpr/WeiRKS16, DBLP:conf/cvpr/0009XLW19, DBLP:conf/cvpr/MoonCL19, DBLP:journals/pr/BenzineLPA21}. Human pose estimation works have been deployed into many applications such as digital human driven. However, to the best of our knowledge, few works attempt to utilize the 3D human pose estimated from images/videos for recommendation. In this paper, we research the topic of leveraging 3D human pose for object effects recommendation in micro-videos, benefiting from the fact that body languages are representative of describing a micro-video's core content.

\textbf{Micro-video recommendation \& video product recommendation} With the wide spread  of micro-videos and their increasing popularity, micro-video recommendation and video product recommendation have attracted numerous attention \cite{DBLP:conf/mm/WeiWN0HC19, DBLP:conf/cikm/LiuLTWNSL21, DBLP:journals/kbs/CaoMRQN20, DBLP:conf/mm/JiangWWGWN20, DBLP:conf/aaai/BouchacourtTN18, DBLP:conf/icmcs/LiuXZC20, DBLP:journals/corr/abs-2110-03902, DBLP:conf/itqm/ZhuHH019, DBLP:conf/icig/JinXH19, DBLP:conf/mm/LiLYCXN19, DBLP:conf/mm/ChenNSK19, DBLP:conf/mm/ChengLWH16, DBLP:journals/tmm/ChengWLH17, DBLP:conf/mm/ZhangLAL0JD20}. The common idea of existing methods is to learn powerful features from images/autos/texts for accurate video-item registration \cite{DBLP:conf/mm/WeiWN0HC19,DBLP:conf/cikm/LiuLTWNSL21, DBLP:conf/kdd/LeiLZ0TLM21, DBLP:journals/access/YangWJ20}. In this work, we propose the new topic of object effects recommendation for micro-videos. Different from video product recommendation, we recommend objects to create a video. We only have the major video content information, human behavior information, and don't have other information such as background and other objects other than humans. Besides, we also build a new dataset named Pose-OBE that is tailored for the new topic, which can be used to benchmark object effects recommendation methods.

\textbf{Disentangled representation learning in recommendation systems} In our work, the very important item-aware implicit prototype learning module is inspired from \emph{disentangled representation learning } \cite{ DBLP:conf/iclr/HigginsMPBGBML17, DBLP:conf/aaai/BouchacourtTN18, DBLP:conf/cvpr/YangLCSHW21}, which is recently popular in recommendation systems and has been widely used from different perspectives. Specifically, there are three main types of disentangled representation learning in recommendation systems: user intention disentanglement \cite{DBLP:conf/nips/MaZ0Y019, DBLP:conf/kdd/MaZYCW020},  information intersection disentanglement \cite{DBLP:conf/recsys/ZhangZHC20}, and specific task disentanglement \cite{DBLP:conf/www/ZhengGLHLJ21,DBLP:conf/www/WangXZWS22}. Our method is most similar to the user intention disentanglement methods. However, in contrast to previous methods that rely on unsupervised decoupling \cite{DBLP:conf/icml/LocatelloBLRGSB19},  we introduce supervised signals for better disentanglement in our work. To our knowledge, we are the first to adopt disentangled representation learning for object effects recommendation in micro-videos.

\vspace{-0.2in}
\section{Method}

\subsection{Problem statement}

Given a micro-video and a dataset of items (candidates of object effects), our goal is to score and rank these items according to the content of the micro-video, hence recommending the most suitable items that should be added to the micro-video referring to the ranking result. During training, suppose we have a set of videos and items, denoted by $\mathcal{V}$ and $\mathcal{I}$, respectively, where $v \in \mathcal{V}$ denotes a video and $i\in \mathcal{I}$ denotes an item. The numbers of videos and items are denoted as $|\mathcal{V}|$ and $|\mathcal{I}|$, respectively.  
An item has $F$ factors, denoted by $i=[f_{i,1}, f_{i,2},\cdots, f_{i,F}]$. Factor $f_{i,F}$ is described by natural language, embedded by pre-trained model BERT \cite{DBLP:conf/naacl/DevlinCLT19}, denoted by $f_{i,F}\in \mathbb{R}^{768}$. A deep network should be trained to learn how to extract features from micro-videos and items respectively and map these features into a shared feature space. Then, during inference, given a video and candidate items, an algorithm should be designed to utilize the output of the trained network to rank items, and finally output a list of recommended items $\{i_1,i_2,\cdots, i_n\}$, where $i_n\in \mathcal{I}$. Note the task only require us to consider the situation when the video is human-centered.



\subsection{Overview}
To achieve our object effects recommendation goal, we propose PoseRec, a novel human pose driven object effects recommendation network. A straight-forward idea to solve the problem is to learn high-level video-level feature vectors and item-level feature vectors from videos and item  factors directly. Then, the recommendation ranking can be obtained by computing the similarities between the video vector and different item vectors. However, we find that directly learning from videos introduces bias, which would significantly limit the recommendation performance. For instance, suppose a girl is dancing and we hope to recommend some effects according to the dance style. If we use an auto-encoder to directly encode the video into a feature vector, it is more likely that the vector would store more information about the backgrounds scene rather than the dance style or body language. Nevertheless, one can dance with the same style in different places with different background. To this end, contradictions arise. To solve the issue, we propose to learn the video content from 3D human poses instead of from the whole video. The intuition behind this design choice is that we believe the body language expressed by the 3D human poses is the core content of the human-centered video, which can provide strong cues of user preference. Fig. \ref{fig_framework} illustrates the holistic design of PoseRec. Our proposed framework is composed of a video side and an item side.

On the video side, we first estimate the human poses of the actor from the micro-video. Suppose the video has $T$ frames. We extract the $T$ frame poses of the actor, denoted by $v=\{p_{v,1}, p_{v,2} ,\cdots, p_{v,T}\}$. Each frame of pose is described by 33 landmarks, and each landmark consists of the 3D joint coordinates $(x, y, z)$ and the $visibility$, denoted by $ p_{v,T}\in \mathbb{R}^{33\times4}$. We adopt BlazePose   \cite{DBLP:journals/corr/abs-2006-10204} to extract the poses.  Then, inspired by STGCN \cite{ DBLP:conf/aaai/YanXL18, DBLP:conf/ijcai/YuYZ18}, we regard the 3D human poses as a spatio-temporal graph. Then, the high-level core content of the video can be learned by a graph convolutional network. In particular, for video $v$ with estimated human poses  $P_{v,:}\in \mathbb{R}^{4\times T\times 33}$, we denote the input of the $l$-th graph convolution layer as $g_v^{l-1}\in \mathbb{R}^{C^l\times T^l\times 33}$, and the output as $g_v^{l}\in \mathbb{R}^{C^{l-1}\times T^{l-1}\times 33}$. $C^l$ is the number of channel after $l$-th graph convolution, and $T^l$ is the length of time dimension after $l$-th graph convolution. The convolutional process can be represented as:
\vspace{-0.05in}
\begin{equation}
\vspace{-0.05in}
\begin{aligned}
g_v^{l}=A g_v^{l-1} W^l
\end{aligned}
\end{equation}

 where $W^l$ is the learned parameter of $l$-th graph convolution layer. $A$ is the normalized graph adjacency matrix.  And then, to get the global graph-level feature instead of only learning node-level local features, we conduct average pooling to obtain $g_v\in \mathbb{R}^{C^L}$, where $C^L$ is the number of channels after the last graph convolution operation. Then, we use a linear transformation operation to map  $g_v$ into a more discriminative presentation $e_v\in \mathbb{R}^{d}$:
 
\vspace{-0.05in}
\begin{equation}
\vspace{-0.05in}
\begin{aligned}
e_v=W_1 g_v+b_1
\end{aligned}
\end{equation}
where $W_1$ and $b_1$ are learned weights and bias respectively.

On the item side, given the factors of each item, e.g., the name, color, shape of the object, etc., we first use a pre-trained BERT \cite{DBLP:conf/naacl/DevlinCLT19} to embed each factor into a factor vector $f_{F}\in \mathbb{R}^{768}$. Then, to integrate all factor vectors of an item into a global item description, we adopt a weighted feature merging strategy. The weight of each factor is a learned parameter. Then, the initial item description $s_i\in \mathbb{R}^{768}$ is the weighted sum of all factor vectors:

\vspace{-0.1in}
\begin{equation}
\vspace{-0.05in}
\begin{aligned}
 s_i=\sum_{j=1}^{F}w_j f_{i,j}
\end{aligned}
\end{equation}
where $w_j$ is learned parameters. $F$ is the number of factors. Finally, we further use a linear transformation operation to map the initial representation to a more discriminative representation $e_i$: 
\vspace{-0.05in}
\begin{equation}
\vspace{-0.05in}
\begin{aligned}
e_i=W_2 s_i+b_2
\end{aligned}
\end{equation}

where $W_2$ and $b_2$ are learned weights and bias respectively. After that, for each video, the corresponding recommendation scores of each item can be obtained by calculating the similarity between $e_i$ and $e_v$:

\vspace{-0.05in}
\begin{equation}
\vspace{-0.05in}
y_{i,v} = sim(e_i, e_v)=\frac{e_i \cdot e_v} {|e_i||e_v|}
\label{eq_sim}
\end{equation}

At inference time, for efficiency, we use the item side to calculate the representation matrix $M$ of all items in advance in an offline manner. Item $i$'s representation $e_i$ is stored in the $i$-th line of $M$. Then, given a video $v$, we can predict the video representation $e_v$  on-the-fly. Finally, the scoring and ranking results can be obtained by: $ y_{:,v} =  e_v\cdot M^T$.

As mentioned before, the human pose driven object effects recommendation task faces two inherent challenges, i.e., issues caused by pose-item registration ambiguity and sparsity. To this end, we propose the item-aware implicit prototype learning module and pose-aware transductive hard-negative mining module to solve the issues. Next, we will introduce the two modules and the loss function we use in detail.

\vspace{-0.1in}
\subsection{Item-aware implicit prototype learning module}

The first issue we hope to solve is the problem caused by ambiguity, which is essentially caused by the diversity and fine-graininess of items. For instance, suppose in a video an actor is playing tennis, and the network should be encouraged to recommend tennis related items such as tennis balls, rackets, sports shoes, sports drinks, etc. However, as far as well know, though they are all highly-related to playing tennis, semantic meanings of objects like drinks and objects like tennis balls are significantly different. Therefore, in the features space, the risk of these objects located far away from each other is very high, which may consequently suffer the recommendation process, especially when we hope to recommendation multiple items with diversity. Inspired by user intention disentanglement \cite{ DBLP:conf/nips/MaZ0Y019}, to solve the issue, we propose the item-aware implicit prototype learning module.

The idea behind this module is that we assume there exist some prototypes that can present some shared characteristics of different items. And we hope each item can be mapped into the prototype space. Our expectation is that these prototypes can implicitly cluster different items into  different groups according to the special characteristics of each prototype holds. And this kind of clustering is irrelevant to the item's category (name) but relevant to the attributes and the role of the item. So, the prototype can link different items and poses together according to their relationship with these characteristics in the prototype space.  For instance, suppose we have two prototypes, one represents the sports-related characteristics, and the other represents edible-related characteristics. Taking playing tennis as an example again, though semantic meanings of drinks and tennis balls are significantly different and they should distribute far away from each other in the normal features space, their distribution in the prototype space could be close due to their special characteristics w.r.t the two prototypes and the video. The only issue is that we should know the rule of mapping items into prototype space. Besides, the rule should be constrained by a specific video. 

To achieve so, we first use $K \times d$ learnable parameters to learn $K$ prototypes ${r_1,r_2, \cdots, r_K},r_K\in \mathbb{R}^d $ during training to consist of the prototype space. To map an item $i$ into the prototype space, we divided the item's representation $e_i$ into $K$ chunks, denoted by $e_i=[e_i^{(1)}: e_i^{(2)}: \cdots: e_i^{(K)} ], e_i\in \mathbb{R}^{K\times d}, e_i^{(K)} \in \mathbb{R}^d$. Each chunk is defined to be related to a prototype. Similarly, given a specific video, the representation of the video can be also divided into $K$ different chunks, denoted by $e_v=[e_v^{(1)}: e_v^{(2)}: \cdots: e_v^{(K)} ], e_v\in \mathbb{R}^{K\times d}, e_v^{(K)} \in \mathbb{R}^d$. Each chunk is also related to a prototype.

The goal is to link $e_v$ and $e_i$ in the prototype space and calculate the recommendation score. To achieve the goal, we first calculate the contribution of each prototype. Specifically, similar as learning $e_i$, for item $i$, we first learn a new item representation $e_{i,c}\in \mathbb{R}^d$. Let $\omega_{i,k}$ be the contribution of  prototype $r_k$ when mapping item $i$ into the features space. We then compute $\omega_{i,k}$ as:

\vspace{-0.1in}
\begin{equation}
\vspace{-0.05in}
\omega’_{i,k}=sim(e_{i,c}, r_k), \omega_{i,:} = softmax(\omega'_{i,:})
\end{equation}

From another respective, $\omega’_{i,k}$ reflects what characteristics item $i$ is mostly relevant to. Using $\omega’_{i,k}$ and the mapping rule, we could map item $i$ into the prototype space. However, as mentioned before, the mapping rule is dynamically constrained by different videos and it is hard to explicitly describe the constraints. So does the mapping rule. Therefore, to ease the task, we regard the rule as a black box and propose to implicitly conduct the mapping by directly calculating the recommendation score using $\omega’_{i,k}$ and chunked $e_v$ and $e_i$. The calculation process can be represented as:
 \vspace{-0.15in}
\begin{equation}
\vspace{-0.05in}
y_{i,v} = \sum_{k=1}^{K}\omega_{i,k} \cdot sim(e_i^{(k)},e_v^{(k)})
\end{equation}

\vspace{-0.15in}
\subsection{Pose-aware transductive hard-negative mining}
To train the network, we use a triplet loss (detailed later) and adopt the hard-negative mining  strategy to help the network learn more discriminative features. Specifically, for each iteration, we sample a batch of videos and their corresponding labeled items for training. For each video, its corresponding labeled items are regarded as positive items, while that of other videos are regarded as negative items. When calculating the loss, for each video, only negative items that have a loss value larger than a threshold are adopted for updating the network parameter. This selection process is called hard-negative mining. In this way, it would be easier for the network to learn better features since the network would be encouraged to focus on pushing way the most dissimilar items in the feature space. In our object effect recommendation task, it is easy to sample positive items, however, it is challenging to obtain hard negative items. This is caused by the fact that different videos can share the same positive items.  We can not simply regard all the attached items of other videos in the batch as negative samples. For example, suppose we have two videos, in one an actor is playing tennis while in the other one an actor is running. In this case, sneakers would be a positive sample of both videos. If the two videos are sampled into a batch, ambiguity would happen, i.e., sneakers will be regarded as both positive items and negative items of a video simultaneously.  We call the phenomenon \emph{sparsity of targeting items}. To solve the issue, we propose the pose-aware transductive hard-negative mining.

In particular, we take advantage of the 3D human poses to solve the problem. Our design is based on the following observation and assumption: if the content of two videos is dissimilar, their corresponding items will also be dissimilar. For example, sneakers would be corresponded to playing tennis but they will never corresponded to cooking. Inspired by this, we propose to first calculate similarities between different video vectors in a batch. If the similarity between two videos is higher than a threshold $p$, we regard them as similar videos, otherwise, they are dissimilar videos. For each video, we randomly select negative items only from the corresponding items of its dissimilar videos for hard negative mining. In another word, 
we mine negative items through transductive pose information interaction. The video vectors learned from human poses act as a proxy for mining hard-negative samples. There are two advantages to the proposed module. First, selecting negative samples from dissimilar videos avoid the risk of adding false negative samples into mining. Second, random sampling ensures the efficiency of the mining process. Note our proposed method is much more efficient than traditional hard negative mining. Complexity analysis can be found in supplementary materials.

\vspace{-0.05in}
\subsection{Loss}

The loss of PoseRec consists of three parts: 
\vspace{-0.1in}
\begin{equation}
\vspace{-0.1in}
L=L_{triple} + L_{pro} + L_{label}
\end{equation}

$L_{label}$ is used to encourage the recommendation score (similarity) between positive video item pairs to be close to 1, otherwise close to 0. We implement the $L_{label}$ as a simple binary cross entropy loss:

\vspace{-0.15in}
\begin{equation}
\vspace{-0.1in}
    \begin{aligned}
L_{label}= &\sum_{v \in \mathcal{B}}\sum_{i \in \mathcal{I}_v}[y_{i,v}'log(sigmoid(y_{i,v})) \\
&+ (1-y_{i,v}')log(1-sigmoid(y_{i,v}))]
    \end{aligned}
\end{equation}

where $B$ is the batch of videos. $y_{i,v}$ is the recommendation score between item $i$ and video $v$,  $y_{i,v}'$ is the ground-truth label between item $i$ and video $v$. $\mathcal{I}_v$ is all the items interacted with $v$.

$L_{pro}$ is designed to encourage different prototypes to distribute as far away with each other as possible in the feature space.
\vspace{-0.15in}
\begin{equation}
\vspace{-0.05in}
L_{pro}=\sum_{k_1=1}^{K-1}\sum_{k_2=k_1+1}^{K}sim(r_{k_1}, r_{k_2})
\end{equation}
where $r_{k_1}$ and $r_{k_2}$ represent  different prototypes.

$L_{triple}$ acts as encouraging the negative samples to distribute far away from the anchor video while encouraging positive samples to distribute close to the anchor. The pose-aware transductive hard-negative mining is adopted to sample negative items.

\vspace{-0.15in}
\begin{equation}
\vspace{-0.05in}
L_{triple}=\sum_{v \in \mathcal{B}}\{\max_{i\in \mathcal{I}_v^{-}}[sim(e_v, e_i)]- \min_{i\in \mathcal{I}_v^{+}}[sim(e_v, e_i)]\}
\end{equation}

where $B$ is the batch of videos. $\mathcal{I}_v^{+}$ is the positive items interacted with $v$, and $\mathcal{I}_v^{-}$ is the mining space of hard-negative mining. 

\begin{table*}[]\centering
\caption{Overall performance on instance recommendation and category recommendation. Best performance is bolded,  and next-best is underlined.}
\vspace{-0.15in}
\label{tab:my-table}
\setlength{\tabcolsep}{0.9mm}{\small{

\begin{tabular}{c|cccccc|cccccc} \hline
          & \multicolumn{6}{c|}{Ins Rec}                   & \multicolumn{6}{c}{Cat Rec}                              \\ \cline{2-13} 
          & R@5    & N@5    & R@10   & N@10   & R@20   & N@20   & R@5    & N@5    & R@10   & N@10   & R@20       & N@20         \\ \hline
Random     & 0.0211 & 0.0212 & 0.0473 & 0.0472 & 0.0819 & 0.0818 & 0.048  & 0.0476 & 0.0876 & 0.0876 & 0.1693     & 0.1702       \\
Pop &
  \textbackslash{} &
  \textbackslash{} &
  \textbackslash{} &
  \textbackslash{} &
  \textbackslash{} &
  \textbackslash{} &
  0.0762 &
  0.0531 &
  0.3753 &
  0.3862 &
  0.3876 &
  0.3952 \\ 
FM         & 0.0497 & 0.0518 & 0.1139 & 0.1209 & 0.1806 & 0.1931 & 0.2688 & 0.2874 & 0.5423 & 0.5999 & 0.6985     & 0.7289       \\
DeepFM     & 0.0194 & 0.0226 & 0.1014 & 0.1126 & 0.1807 & 0.2007 & 0.2365 & 0.2595 & 0.3507 & 0.3803 & 0.5331     & 0.6041       \\
NCF        & 0.0396 & 0.0412 & 0.0748 & 0.0799 & 0.1759 & 0.1963 & 0.2366 & 0.2596 & 0.3451 & 0.3689 & 0.6025     & 0.6734       \\
AFN        & 0.0634 & 0.0719 & 0.0907 & 0.0978 & 0.1985 & 0.2096 & 0.064  & 0.0665 & 0.2949 & 0.319  & 0.6055     & 0.6642       \\
FRNET      & 0.0303 & 0.0386 & 0.1083 & 0.1125 & 0.192  & 0.2108 & 0.0942 & 0.1082 & 0.3351 & 0.3697 & 0.6347     & 0.699        \\ \hline
C3D &  \underline {0.1253} &  \underline {0.1361} &  0.174 &  0.1888 &  0.2717 &  0.2943 &  \underline {0.4718} &  \underline {0.5234} &  \underline {0.6364} &  \underline {0.6844} &  0.7601 &  0.8017 \\
P3D        & 0.113  & 0.1276 & 0.1736 & 0.1977 & 0.2944 & 0.3224 & 0.4651 & 0.5028 & 0.6097 & 0.6576 & 0.7267     & 0.7764       \\
CSN-26 &   0.1107 &  0.118 &  \underline {0.1943} &  \underline {0.2127} &  \underline {0.2989} &  \underline {0.3307} &  0.4285 &  0.4669 &  0.5519 &  0.5986 &  0.7162 &  0.7603 \\
CSN-50     & 0.1039 & 0.1177 & 0.1513 & 0.1718 & 0.2701 & 0.2972 & 0.413  & 0.4503 & 0.5083 & 0.548  & 0.6915     & 0.7306       \\
CSN-101    & 0.0904 & 0.1056 & 0.1551 & 0.1755 & 0.2834 & 0.3061 & 0.3785 & 0.4235 & 0.5642 & 0.6103 & \underline {0.77} & \underline {0.8113} \\
CSN-152    & 0.0638 & 0.0723 & 0.1246 & 0.1433 & 0.2158 & 0.2405 & 0.3591 & 0.4021 & 0.5486 & 0.5967 & 0.7459     & 0.7927       \\ \hline
PoseRec &  \textbf{0.1304} &  \textbf{0.1539} &  \textbf{0.2067} &  \textbf{0.2375} &  \textbf{0.3129} &  \textbf{0.3515} &  \textbf{0.5355} &  
\textbf{0.5904} &  \textbf{0.6444} &  \textbf{0.696} &  \textbf{0.7919} &  \textbf{0.8328}  \\ \hline
\end{tabular}
}}
\vspace{-0.2in}
\end{table*}
\vspace{-0.2in}
\section{Experiment}

We experimentally answer the following questions to evalate the effectiveness of our method: 
\textbf{RQ1}: How does our proposed PoseRec framework perform compared with baseline methods on Pose-OBE? Particularly, for the object effects recommendation task, is utilizing 3D human pose better than directly extracting features from video? \textbf{RQ2}: What does the item-aware implicit prototype learning module actually learn?  \textbf{RQ3}: What is the role of item-aware implicit prototype learning module and pose-aware transductive hard-negative mining in the proposed method?

\textbf{Datasets} To benchmark object effects recommendation methods, we build a novel dataset named Pose-OBE, which consists of
212 micro-videos and 1,087 items (object effects). Each video is annotated with object effects that are most suitable for the scenario, by an operation expert. 

We collect micro-videos from 3 categories (i.e.daily action, sports, art) and 14 subcategories (i.e. eating, football, riding, dancing) human behaviour on TikTok, which have more than 1.4 hours in total. To fit our task, only human-centered videos are considered and others are discarded. For each downloaded micro-video, we manually check it to make sure the whole body of the person in video can be observed and edit each video to delete meaningless frames. 

And we define 1,087 items in advance, including foreground items (i.e. clothes, instruments, sports) and background items (i.e. court, bedroom). These items are always products sold online or visual effects added into videos in post processing. Each item is tagged with a 9-dimensional natural language description including name, usage, shape, color, material,style, color, pattern, and other descriptions. Each description is a short sentence of several words. 

Then, the experts are asked to choose items that are most suitable for video and give each selected item a correlation score ranging from - to 1. Higher scores mean higher correlation between the item and human behavior. Each video is annotated by at least 3 experts. The final score is the average of scores annotated by each expert. Finally, each micro-video corresponds with 3 to 10 items. We randomly divide the dataset into the training set, validation set, and test set in a ratio of 6:2:2. For more details about Pose-OBE, please refer to the supplementary materials.

\textbf{Implementation details} 
We adopt recall at top-k (R@k) and NDCG at top-k (N@k) for evaluating personalized ranking following  \cite{DBLP:conf/nips/MaZ0Y019}. The R@k measures the proportion of positive items in the top-K list to all positive items across all videos. The N@k takes the position of correctly recommended items into account by assigning higher scores to the top hits. For other implementation details, please refer to the supplementary materials.



\begin{figure*}[t]
  \centering
    \subfigure{
		\includegraphics[width=0.35\linewidth]{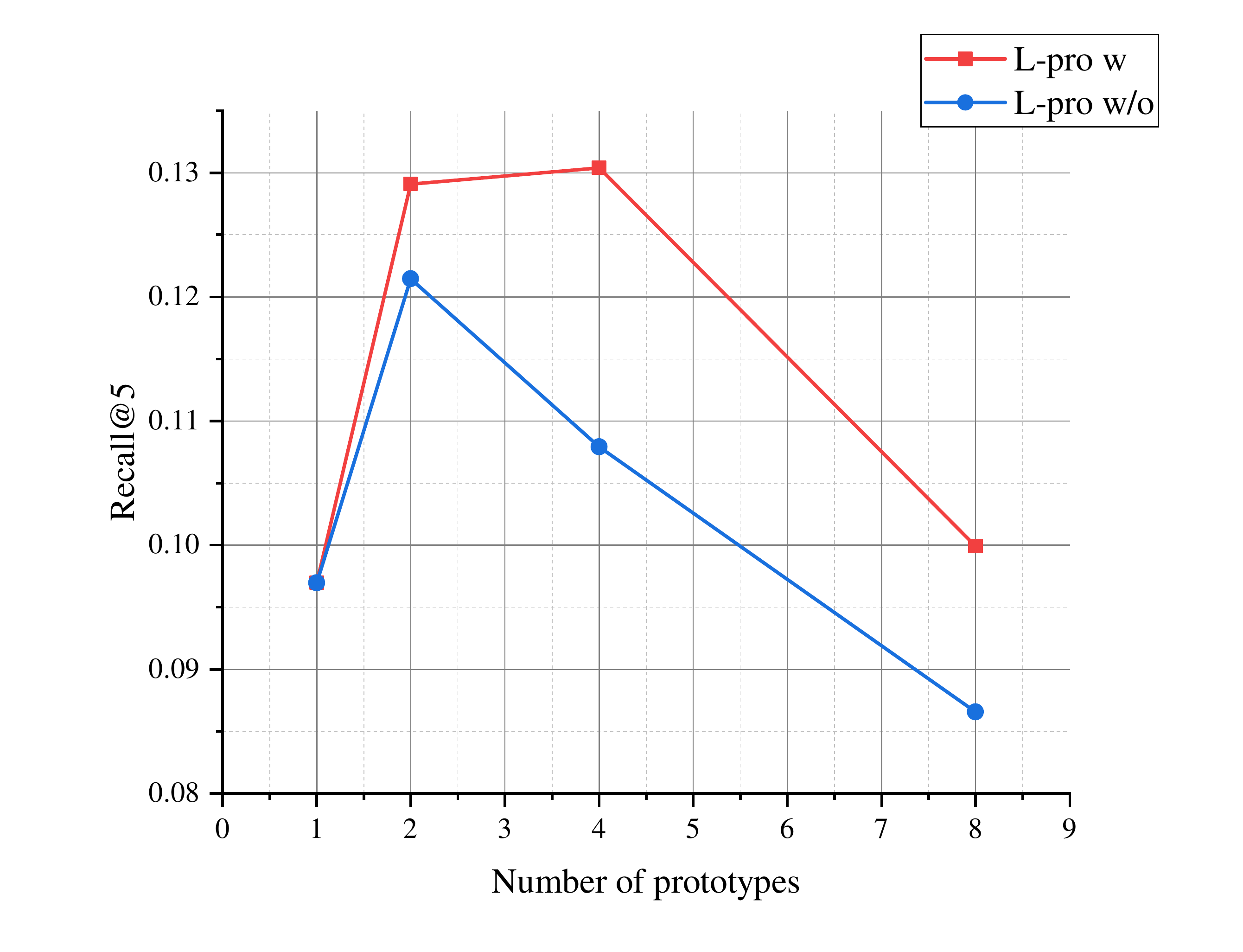}}
	\subfigure{
		\includegraphics[width=0.35\linewidth]{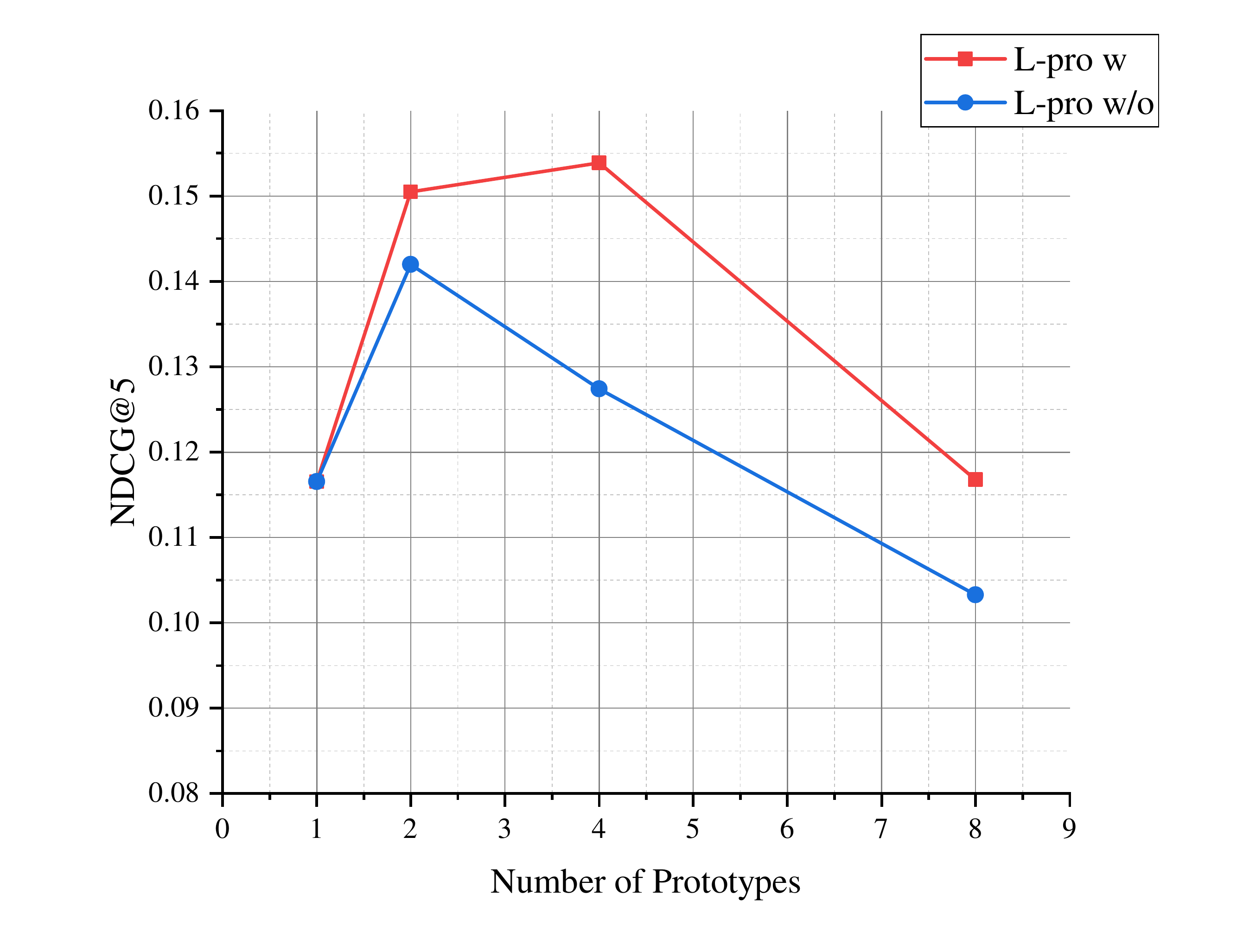}}
		\vspace{-0.15in}
    \caption{Performance comparison (Left: R@5, Right: N@5) of the number of prototypes.}
    \vspace{-0.2in}
	\label{fig_cat}
\end{figure*}

\vspace{-0.15in}
\subsection{Performance comparison (RQ1)}
\vspace{-0.05in}

We perform two kinds of experiments on Pose-OBE, instance-Recommendation (Ins Rec) and Category Recommendation (Cat Rec). Ins Rec treats items with the same item name but different other factors as different items, and calculates metrics at the item level. While Cat Rec regards the items with the same item name as the one category, and the feature vector of one category is calculated as the mean of all item vectors of this category.  The metrics are calculated at the category-level.  Since we are the first to design algorithms for object effects recommendation in micro-videos. There is no existing work for us to compare with. Therefore, we modify some existing recommendation methods to act as strong baselines. Specifically, there are two categories of baselines we compare with. The first category is recommendation methods, including \textbf{Random}, \textbf{Pop}, \textbf{FM}\cite{DBLP:conf/icdm/Rendle10}, \textbf{DeepFM}\cite{DBLP:conf/ijcai/GuoTYLH17},  \textbf{NCF}\cite{DBLP:conf/www/HeLZNHC17}, \textbf{AFN}\cite{DBLP:conf/aaai/ChengSH20}, and \textbf{FRNET}\cite{wang2021enhancing}. The second category is methods that directly learn features from videos, including \textbf{C3D} \cite{DBLP:journals/corr/TranBFTP14}, \textbf{P3D} \cite{DBLP:conf/iccv/QiuYM17}, and \textbf{CSN} \cite{DBLP:conf/iccv/TranWFT19}(a ResNet \cite{DBLP:conf/uemcom/VermaQF17} like network). The comparison result is shown in Table \ref{tab:my-table}. We get the following conclusion from the result: 


 \textbf{First, our proposed PoseRec framework significantly outperforms all strong baselines w.r.t all metrics and all settings }: For example, PoseRec outperforms the next-best baseline C3D with respect to N@5 on Ins Rec by 13\%, and outperforms it with respect to R@5 on Cat Rec also by 13\%. Compared to the "directly feature learning" methods, the superiority of PoseRec demonstrates that learning video content from 3D human poses is a better choice than learning features from the whole image/video. The reason is that body language hidden in human poses is more powerful in representing a human-centered video than extracting features from background scenes.  

\textbf{Second, complex models do not mean better performance}: On the one hand, it is habitual thinking that information contained in an image is richer than the information contained in human poses (33 3D coordinates). However, our work proves that simper data can bring better performance. This may be because that learning information from images is harder than learning information from poses. Besides, in human-centered videos, background scenes in images should be regarded as noise for object effects recommendation. On the other hand, we find that though the network architecture of CSN is more complex than our model, its performance is far behind ours. The reason is that it is much easier for a complex model to get over-fitting. Especially when the background of the images should be regarded as noise.

\begin{table}[t] \centering
\caption{Impact of hard-negative mining}
\vspace{-0.15in}
\label{tab:my-tablertriple}
\setlength{\tabcolsep}{0.9mm}{\small{
\begin{tabular}{c|cc|cc} \hline
             & \multicolumn{2}{c|}{Ins Rec}      & \multicolumn{2}{c}{Cat Rec}      \\ \cline{2-5} 
             &  R@10   & N@10     & R@10   & N@10   \\ \hline
w/o $L_{triple}$ & 0.1094 & 0.1266 &  0.5502 & 0.6122 \\
w $L_{triple}$ &
  \textbf{0.2067} &  \textbf{0.2375} &  \textbf{0.6444} &  \textbf{0.696} \\ \hline
\multicolumn{1}{c|}{$\Delta$} &
  88.94\% &  87.60\% &  17.12\% &  13.69\% \\ \hline
\end{tabular}
}}
\vspace{-0.2in}
\end{table}
\vspace{-0.1in}
\subsection{What does item-aware implicit prototype learning module learns (RQ2)}

In this section, we illustrate what information the item-aware implicit prototype learns to  benefit the object effects recommendation task. In Fig. \ref{fig_sample} (b), we visualize the type embedding using T-SNE. We get two interesting findings. The first finding is that different types of items with different semantic meanings can be well-clustered into different clusters referring to the learned implicit prototypes in the feature space. For example, violin, hat, and basketball are very dissimilar objects, therefore they are distributed in different clusters. The other finding is that we find some items, though share the same name (but with different factors) are distributed differently. We further observe that such kind of distribution is caused by the role of the object. For example, some basketballs are distributed close to the decorations, and some are far away from the decorations. That is because sometimes the basketball is played by an actor, while sometimes the basketball just acts as a decoration on the playground. Hence, basketballs should belong to different prototypes according to their roles. The two findings prove that learning implicit prototypes indeed solves the ambiguity issue because the distribution of items is much more reasonable. In a word, the learned prototypes indeed implicitly present the semantic meanings and roles of items. And this may be one of the main reason why it can help our model improve the recommendation performance.

\subsection{Ablation study (RQ3)}
In this section, we conduct ablation study to investigate the impact of the proposed item-aware implicit prototype learning module and pose-aware transductive hard-negative mining module.
For more ablation studies, please refer to the supplementary materials

\textbf{Impact of item-aware implicit prototype learning} In Fig. \ref{fig_cat}, we study the influence of the number of prototypes $K$ and the influence of $L_{pro}$ in the item-aware implicit prototype learning module. We find: first, using $L_{pro}$ helps the model improve the recommendation performance in all cases except when $K=1$, demonstrating that $L_{pro}$ plays an important role. When $K$ is set to 1, adding $L_{pro}$ makes no sense. Second, with the increase of $K$, the model's performance is increased too, which demonstrates that implicitly learning prototypes helps the model achieve better performance. That is because though this way, the problem caused by ambiguity is solved, so the model can learn better item vectors and video vectors. Third, $K$ shouldn't be too large, otherwise, it would be suffered from over-fitting. Forth, combined with RQ2, $K$ is related to the category of items and more diverse items require a larger $K$. The category information can be used to set $K$.


\textbf{Impact of pose-aware transductive hard-negative mining}
To verify the effectiveness of the  pose-aware transductive hard-negative mining, we remove $L_{triple}$ from the loss function and retrain the network. The result is shown in Table \ref{tab:my-tablertriple}. It can be seen that without $L_{triple}$, the performance of the model significantly drops. This demonstrates that the pose-aware transductive hard-negative mining plays a role in helping the model learn a better shared feature space, hence increasing the recommendation performance. In summary, the proposed pose-aware transductive hard-negative mining is useful and novel.






\vspace{-0.15in}
\section{Conclusion}

In this paper, we research the new topic of object effects recommendation in micro-video platforms. To avoid the problem of introducing background bias caused by directly learning video content from image frames, we propose a network named PoseRec. To overcome problems caused by ambiguity and sparsity, an item-aware implicit prototype learning module and a pose-aware transductive hard-negative mining module are proposed. Besides, a new dataset, Pose-OBE, is tailored for benchmarking object effects recommendation methods is constructed. Extensive experiments on Pose-OBE have demonstrated the superiority of our method. The limitation is that the current algorithm can not process real-time video streams, and we leave it as our future work.

{
\small
\bibliography{aaai23}

}

\appendix

\section{Task and Dataset}

\subsection{Definition of object effects recommendation}

We propose a new topic named object effects recommendation. Here we give the detailed definition of this topic:

In the object effects recommendation task, the input is a micro-video and many pre-defined items. The items are digital objects that can be added into the videos, as special effects. The goal of the task is to learn information from the video to score and rank these items. Then, we can recommend the most suitable items to users or even automatically add the most suitable items to the video via video edit technologies. The object effects recommendation task is widely applicable.

On the one hand, object effects recommendation can help the platform and the video creator to add suitable advertisement according to the video content. The income from advertisement  is  one of the primary incomes of a creator. Therefore, how to help the creator to provide better advertisement is very important for the platform. The income from advertisement is highly-related to the click-through. To encourage audience to click the advertisement, we should make sure that the advertisement we add is what they really interested in. Therefore, we can assume that the more the advertisement is relevant to the video content, the higher the rate the users click the advertisement. Hence, the  object effects recommendation would be very helpful.

On the other hand, object effects recommendation helps creators add visual effects, such as virtual backgrounds, object stickers, and so on, to improve the video quality. Proper effects would significantly increase the quality the micro-video, hence attracting more audience. While impertinent and random effects may lead to a negative impression. For example, we prefer to watch a ballet performance shown at the theatre instead of at the gym. Similarly, when post-processing a ballet show, creators prefer to add a virtual background of theatre instead of the gym.


\subsection{Pose-OBE}

To benchmark object effects recommendation methods, we build a novel dataset named Pose-OBE, which consists of 212 micro-videos and 1,087 items (object effects). Each video is annotated with object effects that are most suitable for the scenario, by a micro-video operation expert. Each item is tagged with a 9-dimensional description including name, usage, shape, color, et al.

To obtain videos, we ask our micro-video operation experts download  micro-videos from Tiktok. To fit our task, only human-centered videos are considered and others are discarded. This ensures that the main subject of the video is a person, so it can be easier to encourage the deep model learn human behaviors for recommendation. For each downloaded micro-video, we   manually check it carefully to make sure the whole body of the person in video can be observed. Unqualified videos are drooped. Besides, we also manually edit each video to delete meaningless frames. Then, we divide all left micro-videos into three categories: daily action, sports, and art. The \emph{daily action} category consists of videos that contain daily life actions,  such as eating, drinking, driving, etc. The \emph{sports} category consists of drastic forceful activities, such as basketball, football, tennis, etc. The \emph{art} category focuses more on artistry, such as dancing, instrument performance, etc.

For each downloaded video, we also ask micro-video operation experts to annotate it with its corresponding items (object effects). Specifically, we first define 1,087 items in advance. These items are always products sold online or visual effects added into micro-videos in post processing.  Each item is tagged with a 9-dimensional natural language description including name, usage, shape, color, material, style, color, pattern, and other descriptions. Each other description is a short sentence of several words. For example, for a chair in a video where a man is playing guitar, the description can be: a chair is used to be a seat, with 1m*0.5m*0.5m in size, with a cylindrical shape, with iron material, with black color, with contract style. Then, the experts are asked to choose items that are most suitable for video and give each selected item a correlation score ranging from - to 1. Higher scores mean higher correlation between the item and human behavior. Each video is annotated by at least 3 experts. We choose the intersection of their selected items as our final results. The final score is the average of scores annotated by each expert. Finally, each micro-video  corresponds with 3 to 10 items.

\begin{table}[] \centering
\caption{The distribution of items}
\label{tab:my-table-item}
\small{
\begin{tabular}{cccc|c|c} \hline
\multicolumn{4}{c|}{Foreground }                 & \multirow{2}{*}{Background} & \multirow{2}{*}{Total} \\ \cline{1-4}
Sports & Instruments & Clothes & Others &                                   &                        \\ \hline
264    & 97          & 101     & 155                    & 470                               & 1087           \\ \hline      
\end{tabular}
}
\end{table}

There are two main categories of items, foreground items, and background items. The foreground items are used or dressed by the actor. They can be subdivided into sports (such as basketball, and football), instruments (such as guitar, and violin), clothes (such as sneakers, and gowns), and other items (such as tableware, and smartphones). The background items do not directly interact with the actor but they are the essential objects for human behavior, such as the football field for football, and the stage for performance. We sorted the items by the above rules and the distribution of items is shown in the Table \ref{tab:my-table-item}.

\section{Complexity analysis of pose-aware transductive hard-negative mining}

Note our proposed method is much more efficient than traditional hard negative mining. We provide the complexity analysis theoretically.  Assuming that the batch size is $b$, and the average numbers of both positive and negative items in each video are both $c$. So a batch consists of $2bc$ items. The mining space of traditional hard-negative mining is $(2bc-c)$, and the complexity of items similarity calculation is $2b^2c$. While the mining space of pose-aware transductive hard-negative mining is $2c$, and the complexity of items similarity calculation is $0.5*b^2+2bc=2bc(0.25b+1)$, which is significantly lower than that of traditional hard negative mining.

\begin{table}[]\centering
\caption{Performance comparison with 2D pose trajectory on Ins Rec}
\label{tab:my-table-2d}
\begin{tabular}{c|cccc}\hline
                   & R@5    & N@5    & R@10   & N@10   \\ \hline
2D human poses & 0.0663 & 0.0729 & 0.1445 & 0.1568 \\
3D human poses & 0.1304 & 0.1539 & 0.2067 & 0.2375 \\ \hline
\end{tabular}
\end{table}

\section{Additional experimental results}

\subsection{Implementation details}

 We use BlazePose \cite{DBLP:journals/corr/abs-2006-10204} to predict 3D human poses from videos.  A sliding time window with a length of 10 and a step of 5 is adopted to divide the micro-videos into 14,281 video samples for training. Each factor of item is embedded with pre-trained BERT \cite{DBLP:conf/naacl/DevlinCLT19} with a hidden size of 768. For all models, we fix the total embedding size $K\times d$ as 256, and $K$ as 4 to guarantee fair comparison. The number of graph convolutional layers is 3 on the video side and other settings are following \cite{DBLP:conf/aaai/YanXL18}. The learning rate is 1e-4. The network is implemented using Pytorch and optimized using the Adam\cite{DBLP:journals/corr/KingmaB14} optimizer. L2 regularization is added to the loss function to avoid over-fitting. All experiments are conducted o a single NVIDIA 3090 GPU.

\subsection{Study on the information source}

\begin{table}[] \centering
\caption{Performance comparison with frames for video embedding}
\label{tab:my-frames}
\begin{tabular}{c|cccc|c} \hline
Frames & R@5    & N@5    & R@20   & N@20   & Time   \\ \hline
10     & 0.1304 & 0.1539 & 0.3129 & 0.3515 & 1      \\
25     & 0.1267 & 0.1478 & 0.3296 & 0.3674 & 1.5454 \\
50     & 0.1319 & 0.1561 & 0.3411 & 0.3836 & 2.6004 \\ \hline
\end{tabular}
\end{table}

\begin{figure*}[]
  \centering
    \subfigure{
		\includegraphics[width=0.4\linewidth]{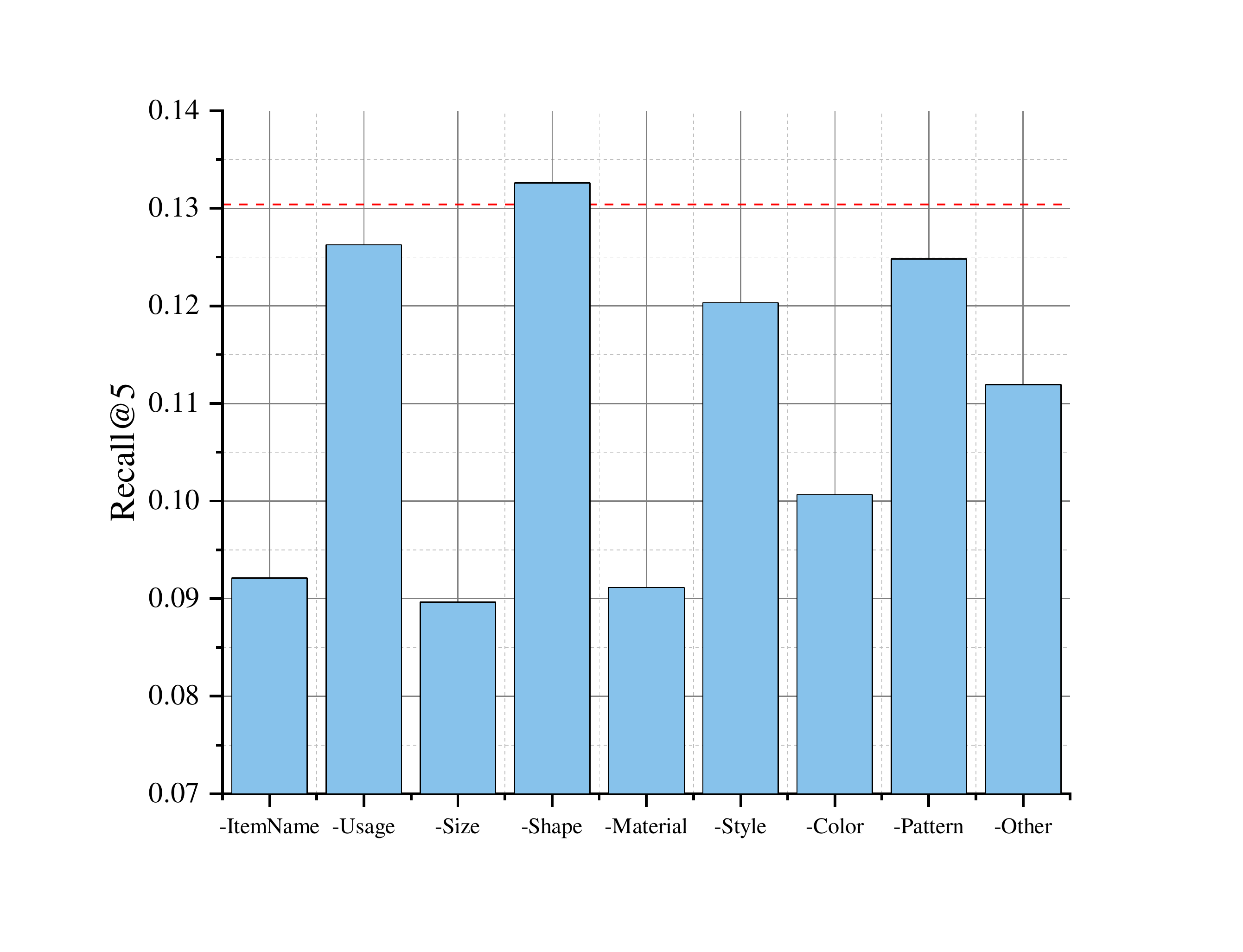}}
	\subfigure{
		\includegraphics[width=0.4\linewidth]{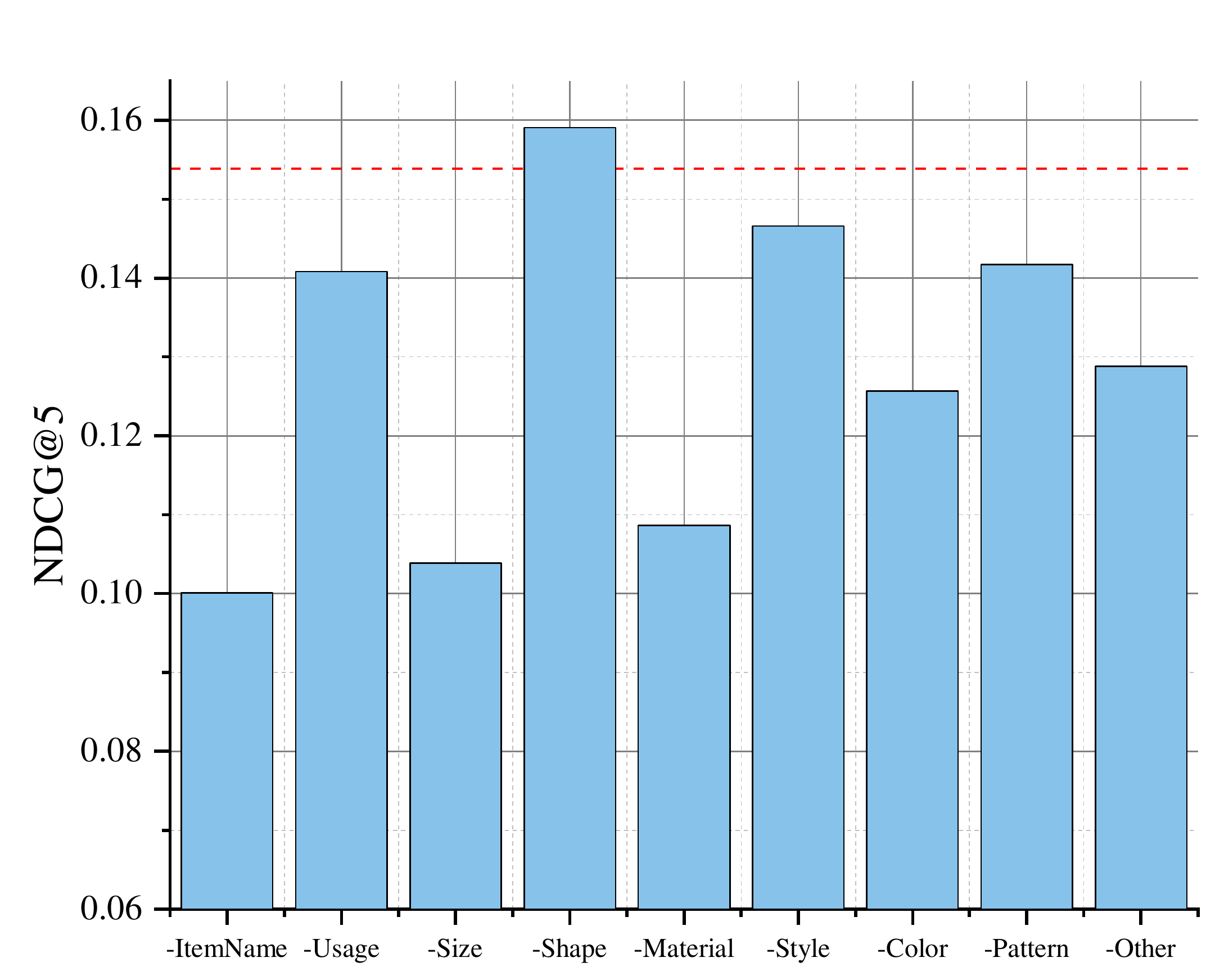}}
    \caption{Effect of different factor on Ins Rec (Left: R@5. Right: N@5). “-” indicates that remove the corresponding factor. The red line indicates that no factor has been removed.}
	\vspace{-0.1in}
	\label{fig_nlp}
\end{figure*}

\begin{figure*}[]
  \centering
    \subfigure{
		\includegraphics[width=0.4\linewidth]{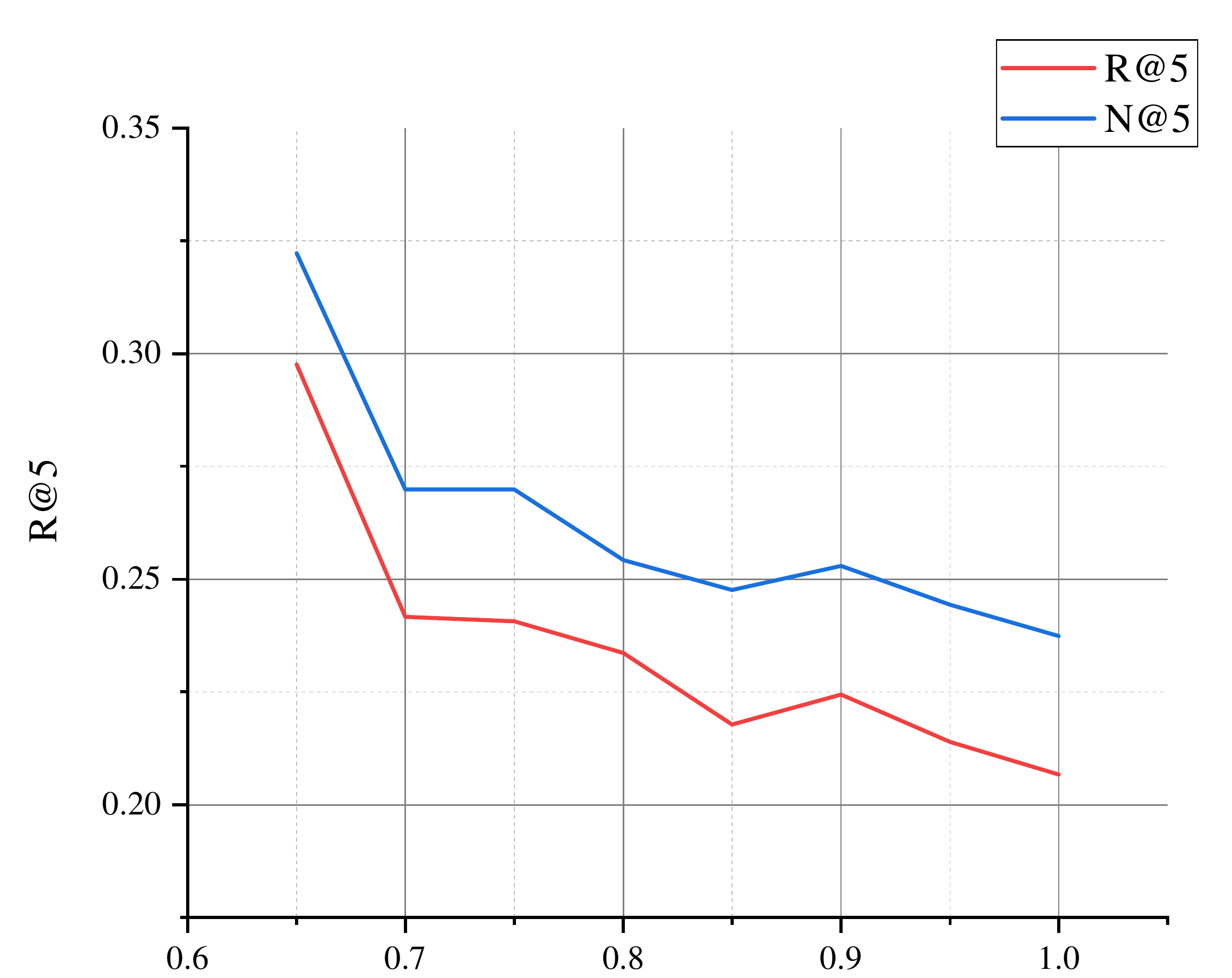}}
	\subfigure{
		\includegraphics[width=0.4\linewidth]{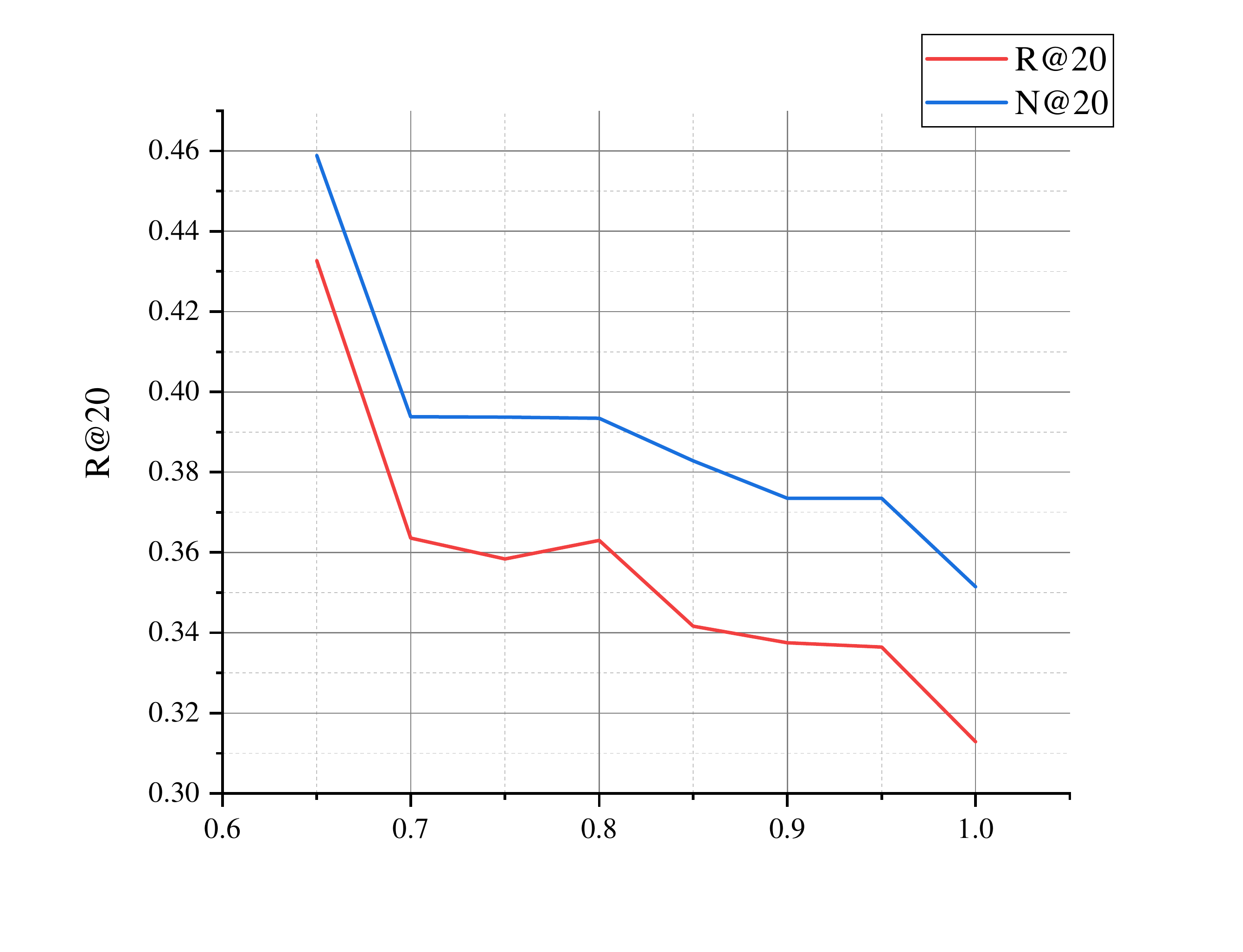}}
    \caption{Effect of the numbers of items on Ins Rec (Left: @5. Right: @20). The horizontal axis of each represents the percentage of the retained item to all. As the percentage increases, the number of items increases.}
	\vspace{-0.1in}
	\label{fig_no}
\end{figure*}

\paragraph{3D human pose or 2D human pose}

In our paper, we have demonstrated that using 3D human poses for object effects recommendation is superior than directly learning features from videos, thanks to the powerful body languages hidden in human poses. Yet, another question arises: should we use 3D human poses? Can we use 2D human poses instead? To answer the question, we conduct experiments and the results are shown  in  Table \ref{tab:my-table-2d}. We find that using 3D human poses significantly outperforms using 2D poses. One possible reason is that 3D human poses contains more information about the body language. Besides, 2D human poses may suffer from problems such as self-occlusion.

\paragraph{Impact of frames for video embedding}
To find how frames for video embedding impact the performance, we vary the input frames for video embedding and the results are shown in Table \ref{tab:my-frames}. More frames may provide more information. But meanwhile, it takes more time and needs a more complicated network to deal with more frames. Thus we balance the two aspects and set the number of frames as 10.

\paragraph{Impact of item factors}

In this part, we study the effect of different factors for each item. In Fig. \ref{fig_nlp}, we study the contribution of different factors by removing one factor each time. We find that the most important factor is \emph{item name, size, and material}. There are also misleading factors, such as \emph{shapes}. It may be caused by the difficulty of annotating deformable objects like cloth as well as the difficulty of describing the shape of an object.

\paragraph{Impact of the number of items}

To study how the number of items affects the performance, we randomly drop the items of the dataset with different probability and the results are shown in Fig. \ref{fig_no}. The horizontal axis of each represents the percentage of the retained item. As the number of objects increases, the performance decreases. It indicates that the growing item space brings difficulty for recommendation.

\subsection{Study on the network structure}

\begin{table}[] \centering
\caption{The structure of different graph convolution layers}
\label{tab:my-table-gcn}
\begin{tabular}{c|cc|cc|cc|cc}\hline
 & \multicolumn{2}{c|}{1-Layer} & \multicolumn{2}{c|}{2-Layer} & \multicolumn{2}{c|}{3-Layer} & \multicolumn{2}{c}{4-Layer} \\ \cline{2-9} 
      & $C^l$ & $T^l$ & $C^l$ & $T^l$ & $C^l$ & $T^l$ & $C^l$ & $T^l$ \\ \hline
$l$=0 & 4     & 10    & 4     & 10    & 4     & 10    & 4     & 10    \\
$l$=1 & 256   & 10    & 64    & 10    & 64    & 10    & 64    & 10    \\
$l$=2 &       &       & 256   & 5     & 128   & 5     & 128   & 5     \\
$l$=3 &       &       &       &       & 256   & 5     & 256   & 5     \\
$l$=4 &       &       &       &       &       &       & 256   & 5    
\\ \hline
\end{tabular}
\vspace{-0.2in}
\end{table}

\begin{figure*}[]
  \centering
    \subfigure{
		\includegraphics[width=0.4\linewidth]{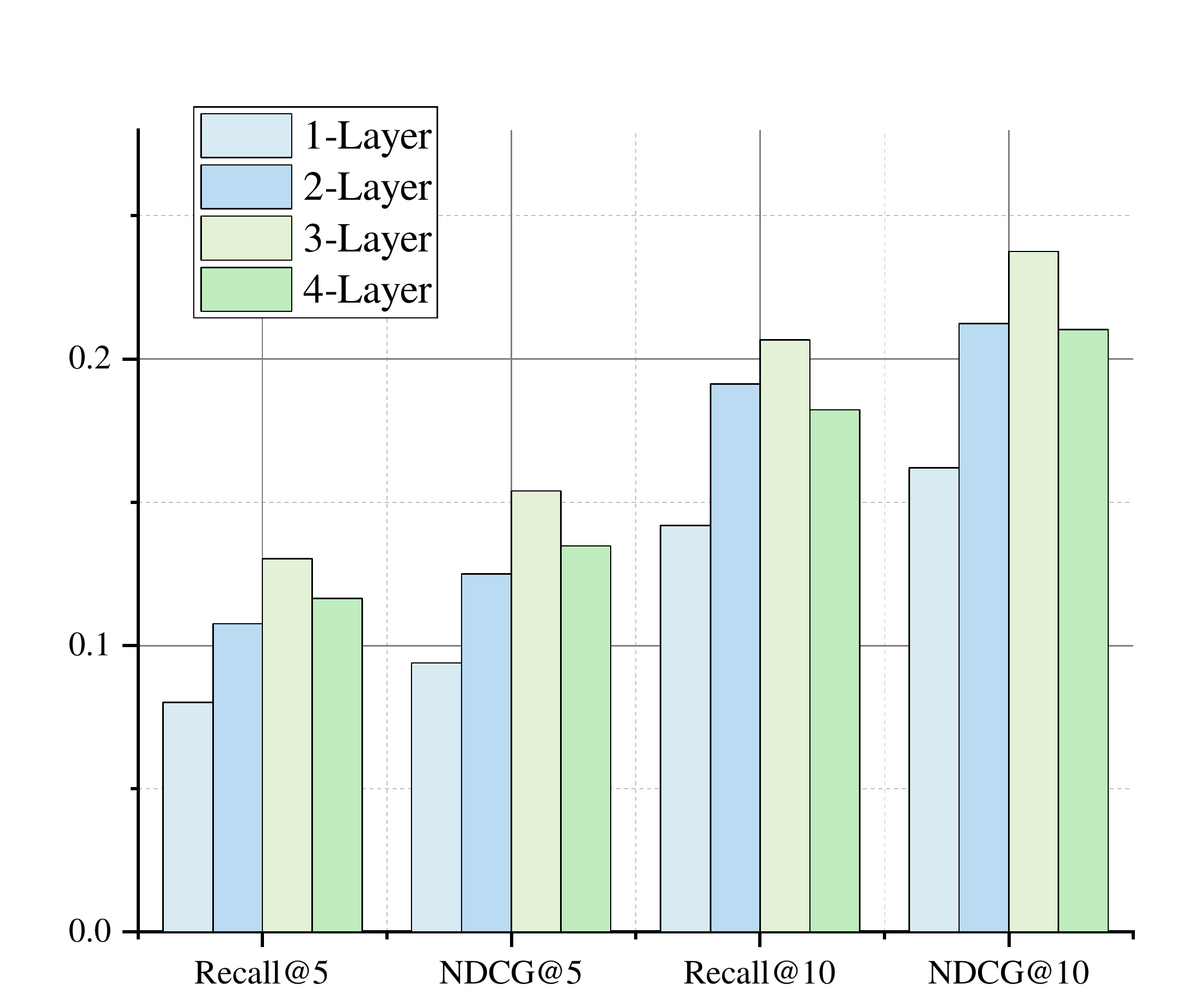}}
	\subfigure{
		\includegraphics[width=0.4\linewidth]{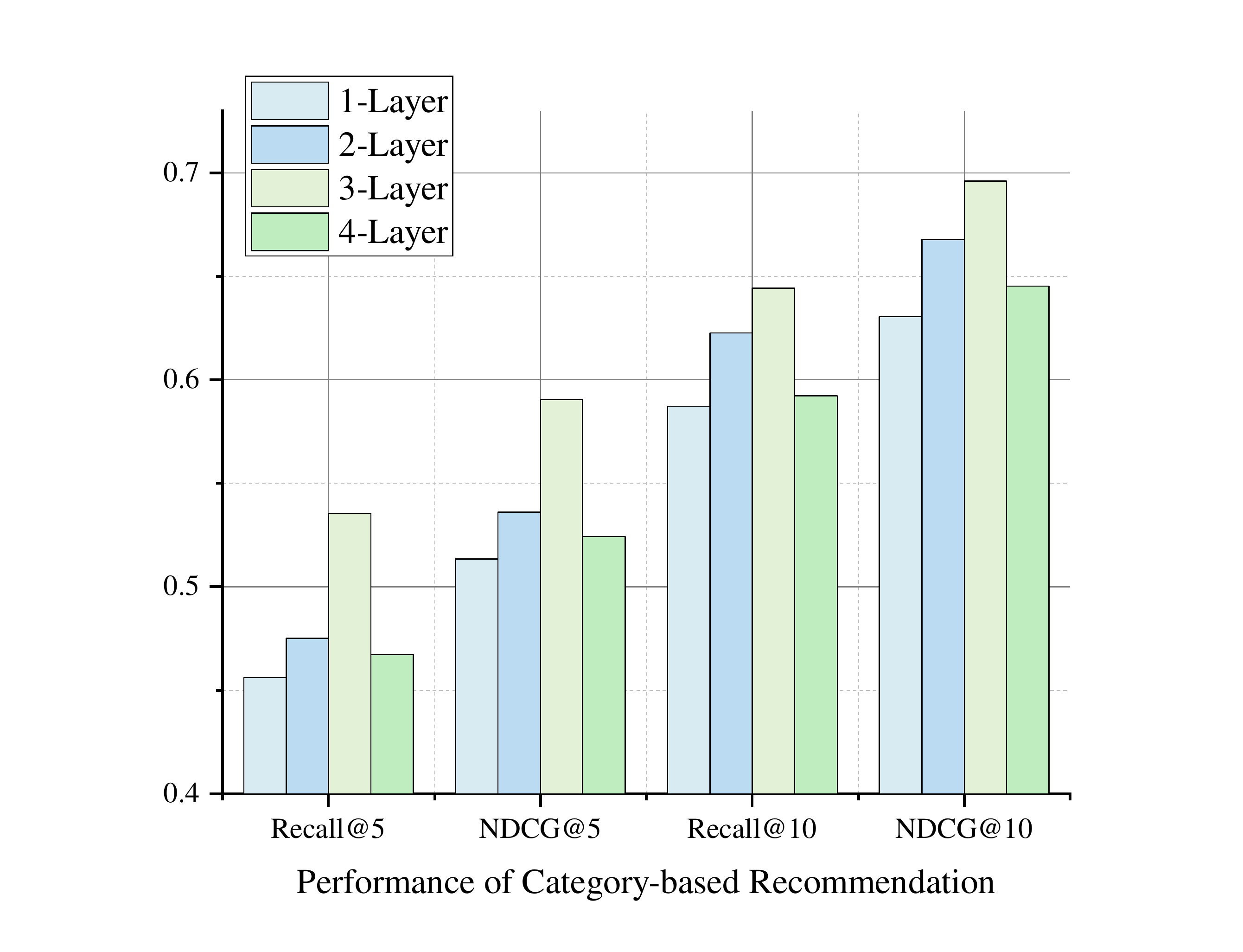}}
    \caption{Performance comparison of the number of GCN layers. Left: Ins Rec. Right: Cat Rec}
	\vspace{-0.1in}
	\label{fig_gcn}
\end{figure*}


\paragraph{Impact of graph convolution layers}

The number of graph convolution layers $L$ on the video side is also very important for the model's performance. We implement networks with different $L$ and the structure has shown in Table \ref{tab:my-table-gcn}. We compare their performance in Fig. \ref{fig_gcn}. It can be seen that $L$ shouldn't be too small nor too large. When it is too small, the network can not learn high level features; when it is too large, it is easy to cause over-fitting. Setting $L$ as 3 is the best choice.




\begin{figure*}[]
  \centering
        \includegraphics[width=0.8\linewidth]{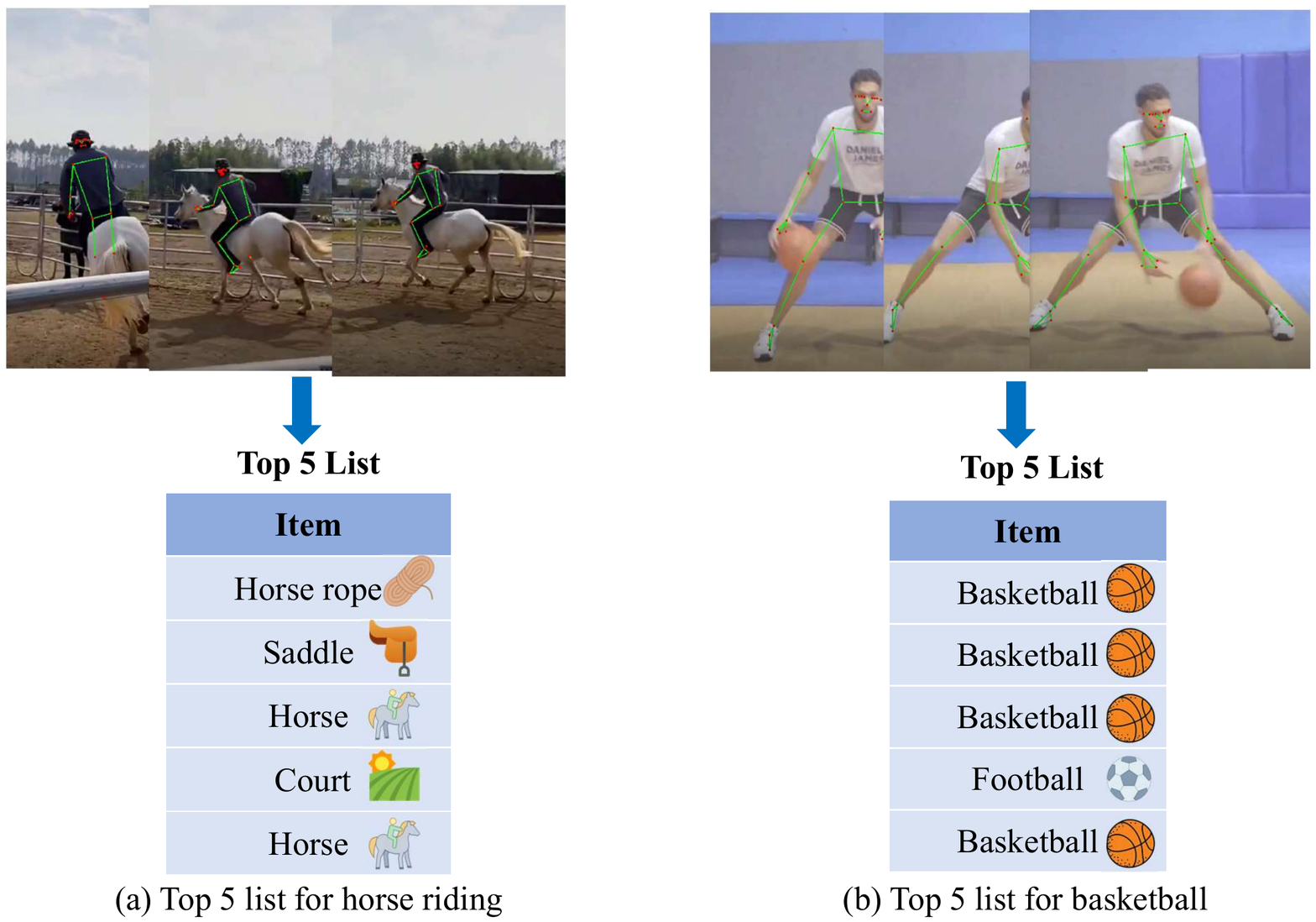}
    \caption{Recommendation visualization for horseback riding (a) and basketball (b)}
	\vspace{-0.1in}
	\label{fig_vis}
\end{figure*}
\subsection{Visualization of recommendation results}
In in Fig. \ref{fig_vis}, we qualitatively show two examples of the recommendation results.  The two videos are randomly selected from the test set, one is  about a man playing basketball, while  another is about a man riding horseback. We show the top 5 recommendation results. We can find from the figure that our methods can recommend objects that highly relevant to the 3D human poses, i.e., the human behavior. Note that we only use the 3D human poses for learning video content. Therefore, though some recommended objects appears in the video, their information is not fed into the network for learning features. 


\end{document}